\documentclass[cameraready]{Interspeech}
\usepackage{adjustbox}
\usepackage{multirow}
\usepackage{dsfont}
\usepackage{pifont}

\title{SFL-MTSC: Leveraging Semantic Frame-Level Multi-Task Self-Consistency for Robust Multi-Intent Spoken Language Understanding}

\author[affiliation={1}, orcid=0009-0006-6816-4943]{Po-Yen}{Chen}
\author[affiliation={1}, orcid=0000-0003-0693-8932]{Berlin}{Chen}

\address{
    $^1$ National Taiwan Normal University, Taiwan
}

\email{cby931001@gmail.com, berlin@ntnu.edu.tw}


\newcommand{\blue}[1]{\textcolor{blue}{#1}}
\newcommand{\red}[1]{\textcolor{red}{#1}}
\usepackage{comment}

\begin{document}

\maketitle

\begin{abstract}
    Prompt-based spoken language understanding (SLU) with large 
    language models (LLMs) often suffers from inconsistent intent--slot structures due to decoding stochasticity, particularly in multi-intent scenarios. In view of this, we propose Semantic Frame-Level Multi-Task Self-Consistency (SFL-MTSC), a novel structured aggregation framework operating at the semantic frame level. Instead of output-level majority voting, SFL-MTSC decomposes predictions into intent-specific frames, applies domain--intent grouping and slot-level clustering, and evaluates cluster reliability using path support scoring. Reliable frames are retained and re-integrated to form the final prediction. Zero-shot experiments on the MAC-SLU benchmark dataset show improved slot F1 and overall accuracy over single-path inference, while intent accuracy remains largely stable across most settings.
\end{abstract}

\section{Introduction}
Spoken Language Understanding (SLU) is a fundamental paradigm for 
task-oriented spoken semantics extraction, widely applied in scenarios 
such as smart homes and in-vehicle systems to interpret user spoken 
commands for executing downstream tasks~\cite{tur2011spoken}. It 
typically consists of two core subtasks: intent detection and slot 
filling~\cite{qinSurveySpokenLanguage2021}. For example, given the 
utterance ``\textit{I want to know the weather in Taipei}'', intent 
detection identifies the overall semantic intent (e.g., 
\texttt{GetWeather}), while slot filling extracts the associated 
arguments (e.g., \texttt{city\_name} = \texttt{Taipei}). Recently, with the advancement of large 
language models (LLMs), prompt-based methods relying on LLMs have emerged as a 
promising paradigm for SLU, enabling zero-shot or few-shot inference 
without task-specific fine-tuning~\cite{panPreliminaryEvaluationChatGPT2023, gpt-slu, croprompt, xingDXANetDualTaskCrossLingual2026, liHowChatGPTRobust, 10447804, aroraIntentDetectionAge2024, Gao, Lib}.

However, prompt-based SLU with LLMs often suffers from inconsistent 
intent--slot structures due to decoding stochasticity, particularly 
in multi-intent scenarios where a single utterance may simultaneously 
express multiple intents across different domains~\cite{macslu}. In such cases, 
different decoding runs may produce conflicting semantic frames, 
leading to unreliable predictions that are difficult to aggregate 
at the output level.

Self-consistency~\cite{self-consistency, taubenfeldConfidenceImprovesSelfConsistency2025,
chenUniversalSelfConsistencyLarge, Ahmed, Nowak} has been proposed to improve LLM reasoning by sampling multiple paths and aggregating via majority voting. While prior work has explored its application to SLU~\cite{croprompt, HIT-SCIR}, these methods were not designed for multi-intent settings with semantic frame-structured predictions. LLM-as-a-judge~\cite{gu2026survey, li-etal-2025-generation} offers an alternative, but requires additional LLM calls, carries hallucination risk~\cite{limitationofLLMasajudge}, 
and operates at the output level, making fine-grained frame-level filtering infeasible.

In this work, we propose \textbf{Semantic Frame-Level Multi-Task 
Self-Consistency (SFL-MTSC)}, a structured aggregation framework 
designed for robust multi-intent SLU. Instead of output-level majority 
voting, SFL-MTSC decomposes predictions into intent-specific semantic 
frames, applies domain--intent clustering and slot-level clustering, and 
evaluates cluster reliability using path support inspired by association rule mining \cite{zhaoAssociationRuleMining2003, ref1}. 
Reliable frames are retained and re-integrated to form the final 
prediction. Our main contributions are as follows:

\begin{itemize}
    \item We propose SFL-MTSC, a frame-level self-consistency 
    framework for multi-intent SLU. It works below the output level, 
    which allows it to better remove false intents and noisy slot 
    predictions.
    
    \item We introduce Hybrid Jaccard similarity for slot clustering. 
    It combines key-based and value-based matching, making the 
    framework more robust when slot key names vary across reasoning 
    paths.
    
    \item We conduct zero-shot experiments on MAC-SLU~\cite{macslu}, a challenging 
    Chinese multi-intent SLU benchmark, across three model 
    configurations spanning text-only LLMs~\cite{yang2025qwen3technicalreport}, ASR+LLM pipelines~\cite{radford2023robust, yang2025qwen3technicalreport}, and 
    end-to-end large audio-language models (LALMs)~\cite{xu2025qwen25omnitechnicalreport}, demonstrating 
    consistent improvements in overall accuracy and Slot F1.
\end{itemize}
All of code and data used in this work will be publicly available at \texttt{\href{https://github.com/boyan1001/SFL-MTSC}{https://github.com/boyan1001/SFL-MTSC}}.

\section{Proposed Approach}
\begin{figure*}[t]
    \centering
    \includegraphics[width=1 \linewidth]{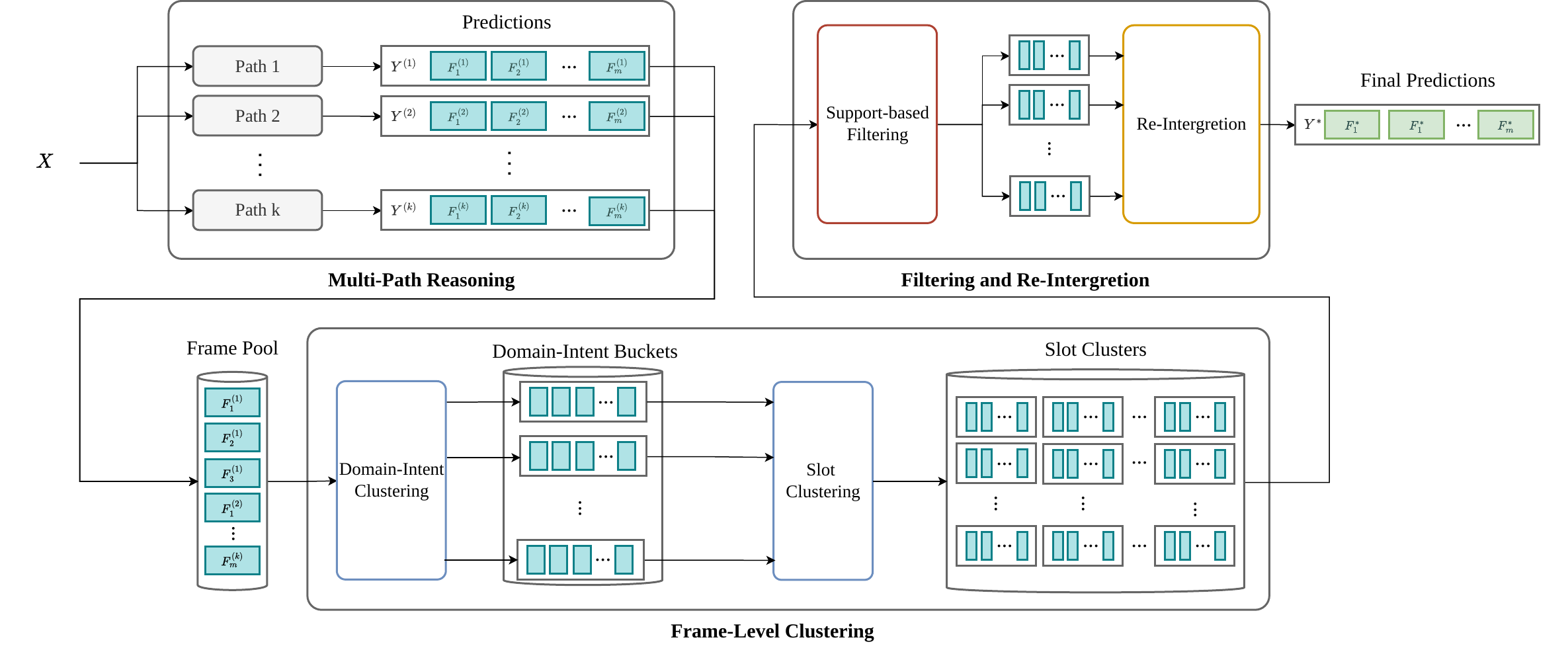}
    \caption{Overview of the SFL-MTSC framework. Given an input utterance $X$, 
multiple reasoning paths are sampled to construct a frame pool. 
Frames are then grouped via domain--intent clustering and slot-level 
clustering, filtered by path support, and re-integrated into the 
final multi-intent prediction $Y^*$.}
    \label{fig:SFL_MTSC_architecture}
\end{figure*}

In this section, we illustrate the workflow of SFL-MTSC as Figure \ref{fig:SFL_MTSC_architecture}. Unlike standard 
majority voting, which operates on the final output level, 
SFL-MTSC decomposes each reasoning path into intent-specific 
semantic frames and evaluates structural consistency at the 
frame level. This allows the framework to selectively retain 
stable frames supported by multiple reasoning paths while 
discarding hallucinated or structurally conflicting ones, 
thereby producing more reliable multi-intent predictions.

\subsection{Multi-Paths Reasoning}
\label{sec:reasoning}

Given an input utterance $X$, we perform inference over the same
LLM or LALM using $K$ distinct reasoning paths, each differing
in sampling temperature. Each semantic frame $F$ is a structured
triple consisting of a domain $d$, an intent $i$, and a set of
slot--value pairs $s$:
\begin{equation}
    F = (d, i, s), \quad s = \{(k_1, v_1), (k_2, v_2), \dots\}
\end{equation}
The prediction of path $k$ consists of a set of such frames:
\begin{equation}
    Y^{(k)} = \{ F^{(k)}_1, F^{(k)}_2, \dots, F^{(k)}_{m_k} \}
\end{equation}
where $m_k$ is the number of frames predicted by path $k$. Due
to model stochasticity, different paths may produce conflicting
intent--slot structures for the same utterance, particularly in
multi-intent scenarios. The subsequent stages of SFL-MTSC are
designed to aggregate these diverse predictions and resolve such
inconsistencies at the frame level.

\subsection{Frame Pool Construction}
\label{sec:pool}

All semantic frames across $K$ reasoning paths are collected 
into a unified frame pool:
\begin{equation}
    \mathcal{F} = \bigcup_{k=1}^{K} Y^{(k)} = \left\{ F_j^{(k)} 
    \mid k \in \{1,\dots,K\},\ j \in \{1,\dots,m_k\} \right\}
\end{equation}
Frames with missing \texttt{domain} or \texttt{intent} fields 
are discarded prior to pooling. Each retained frame is annotated 
with its source path index $p$ for use in the subsequent 
self-consistency scoring stage.

\subsection{Frame-Level Clustering}
\label{sec:clustering}

With $\mathcal{F}$ constructed, we identify frames referring 
to the same underlying semantic intent via a coarse-to-fine 
clustering procedure.

\subsubsection{Domain--Intent Clustering}

We first partition $\mathcal{F}$ into buckets by
\texttt{(domain, intent)} pair:
\begin{equation}
    \mathcal{F}_{d,i} = \left\{ F_j^{(k)} \in \mathcal{F}
    \mid d_j^{(k)} = d,\ i_j^{(k)} = i \right\}
\end{equation}
This prevents frames with different intents from being merged 
and narrows the search space for slot-level comparison. In 
multi-intent scenarios, the same intent label may appear in 
multiple frames within a single path, motivating the need for 
further slot-level clustering within each bucket.

\subsubsection{Slot Clustering}
Within each bucket $\mathcal{F}_{d,i}$, we cluster frames by 
slot similarity using a threshold similarity graph whose connected 
components are taken as clusters, inspired by 
\cite{widdowsGraphModelUnsupervised2002}. The similarity between 
frames is measured by Hybrid Jaccard, inspired by the Jaccard 
index~\cite{jaccard1901etude}, which interpolates Key--Value 
Jaccard $\text{sim}_{kv}$ and Value-Based Jaccard $\text{sim}_{val}$ 
with coefficient $\alpha \in [0,1]$, where $v(s)$ denotes the set 
of all values in slot set $s$.
\begin{equation}
    \text{sim}_{kv}(F_a, F_b) = \frac{|s_a \cap s_b|}{|s_a \cup s_b|}, \quad
    \text{sim}_{val}(F_a, F_b) = \frac{|v(s_a) \cap v(s_b)|}{|v(s_a) \cup v(s_b)|}
\end{equation}
\begin{equation}
    \text{sim}_{hyb}(F_a, F_b) = \alpha \cdot \text{sim}_{kv}(F_a, F_b) 
    + (1-\alpha) \cdot \text{sim}_{val}(F_a, F_b)
\end{equation}
$\text{sim}_{kv}$ is structurally precise but sensitive to key 
naming variation; $\text{sim}_{val}$ is more robust but may 
conflate frames with coincidentally overlapping values. Hybrid 
Jaccard balances both. We then construct a threshold similarity 
graph $G_{d,i} = (\mathcal{F}_{d,i}, E_{d,i})$:
\begin{equation}
    E_{d,i} = \left\{ (F_a, F_b) \mid F_a, F_b \in 
    \mathcal{F}_{d,i},\ \text{sim}_{hyb}(F_a, F_b) \geq \tau \right\}
\end{equation}
The connected components of $G_{d,i}$ are taken as the
final slot clusters, yielding the overall set of
frame-level clusters:
\begin{equation}
    \mathcal{C}_{d,i} = \mathrm{CC}(G_{d,i}), \quad
    \mathcal{C} = \bigcup_{(d,i)} \mathcal{C}_{d,i}
\end{equation}
where each $C \subseteq \mathcal{F}_{d,i}$ represents a distinct 
semantic frame instance.

\begin{table*}[t]
    \centering
    \resizebox{\textwidth}{!}{%
    \begin{tabular}{l l l l l l l l l l}
        \toprule
        \multirow{5}{*}{\textbf{Methods}}&
        \multicolumn{3}{c}{\textbf{Text-NLU}} &
        \multicolumn{3}{c}{\textbf{ASR + NLU (Pipeline)}} &
        \multicolumn{3}{c}{\textbf{End-to-End}} \\
        \cmidrule(lr){2-4} \cmidrule(lr){5-7} \cmidrule(lr){8-10}
        & \multicolumn{3}{c}{\textbf{Qwen3-4B-Instruct-2507}} &
        \multicolumn{3}{c}{\textbf{Whisper + Qwen3-4B-Instruct-2507}} &
        \multicolumn{3}{c}{\textbf{Qwen2.5-Omni-7B}} \\
        \cmidrule(lr){2-4} \cmidrule(lr){5-7} \cmidrule(lr){8-10}
        & \textbf{Overall Acc.} & \textbf{Intent Acc.} & \textbf{Slot F1} 
        & \textbf{Overall Acc.} & \textbf{Intent Acc.} & \textbf{Slot F1}
        & \textbf{Overall Acc.} & \textbf{Intent Acc.} & \textbf{Slot F1}\\
        \midrule

        \textit{Vanilla Prompting \cite{macslu}} & 2.07 & \textbf{46.84} & 29.39 & 2.43 & \textbf{43.63} & 22.15 & 0.64 & \textbf{21.46} & \textbf{4.86} \\
        \textit{+SFL-MTSC} & \textbf{3.30} \blue{(+1.23)} & 44.22 \red{(-2.62)}& \textbf{58.25} \blue{(+28.86)} & \textbf{3.39} \blue{(+0.4)}&  38.31 \red{(-5.32)}& \textbf{49.53} \blue{(+27.38)} & \textbf{2.09} \blue{(+1.45)}& 13.90 \red{(-7.56)}& 4.57 \red{(-0.29)}\\
        \midrule
        
        \textit{CroPropmt (Intent$\rightarrow$Slot) \cite{croprompt}} & 1.51 & \textbf{48.84} & 36.93 & 1.81 & \textbf{43.68} & 23.77 & 0.42 & \textbf{25.30} & \textbf{5.46} \\
        \textit{+SFL-MTSC} & \textbf{1.91} \blue{(+0.40)} & 46.13 \red{(-2.71)} & \textbf{46.24} \blue{(+9.31)} & \textbf{2.17} \blue{(+0.36)}& 41.44 \red{(-2.24)}& \textbf{34.48} \blue{(+10.71)} & \textbf{2.17} \blue{(+1.75)}& 18.59 \red{(-6.71)}& 4.79 \red{(-0.67)}\\
        
        \midrule
        \textit{CroPropmt (Slot$\rightarrow$Intent) \cite{croprompt}} & 4.16 & \textbf{55.12} & \textbf{52.70}  & 4.49 & \textbf{50.86} & 39.01 & 0.94 & \textbf{12.14} & 0.37 \\
        \textit{+SFL-MTSC}  & \textbf{4.52} \blue{(+0.36)}& 53.26 \red{(-1.86)}& 52.23 \red{(-0.47)} & \textbf{4.69} \blue{(+0.20)}& 49.61 \red{(-1.25)}& \textbf{39.54} \blue{(+0.53)} & \textbf{2.43} \blue{(+1.49)}& 6.78 \red{(-5.36)}& \textbf{3.93} \blue{(+3.56)}\\
        
        \midrule
        \textit{GPT-SLU \cite{gpt-slu}} & 4.07 & \textbf{58.05} & \textbf{49.06} & 4.23 & \textbf{53.52} & \textbf{43.16}  & 0.18 & \textbf{14.12} & \textbf{11.24} \\
        \textit{+SFL-MTSC} & \textbf{4.10} \blue{(+0.03)}& 56.82 \red{(-1.23)}& 47.33 \red{(-1.73)} & \textbf{4.26} \blue{(+0.03)}& 50.13 \red{(-3.39)}& 40.02 \red{(-3.14)} & \textbf{2.13} \blue{(+1.32)}& 8.16 \red{(-5.96)} & 3.91 \red{(-7.33)}\\
        \bottomrule
    \end{tabular}%
    }
    \caption{Main results for LLMs and LALMs on MAC-SLU dataset by using different temperature and using SFL-MTSC for each prompting methods. For \textit{Vanilla Prompting}, we follow MAC-SLU methods \cite{macslu} to directly utilize a sample single-round prompting method for SLU. Performance gains/drops compared to  baseline are highlighted with \blue{blue}/\red{red}. The best results are shown in \textbf{bold}.}
    \label{tab:main_result}
\end{table*}
\subsection{Path Support Scoring}
\label{sec:scoring}
After clustering, each cluster $C \in \mathcal{C}$ represents 
a candidate semantic frame instance.  However, clusters vary 
in reliability: a single-path cluster may reflect a 
hallucinated intent, while one with conflicting slot values 
indicates structural inconsistency. Inspired by the support 
measure in association rule mining \cite{zhaoAssociationRuleMining2003, ref1}, 
we quantify cluster reliability by counting the number of 
distinct reasoning paths that contribute frames to the cluster:

\begin{equation}
    \text{sup}_p(C) = \left| \{ p \mid \exists\, F_j^{(p)} \in C \} \right|
\end{equation}

Unlike response probability in CISC~\cite{taubenfeldConfidenceImprovesSelfConsistency2025}, 
which scores each reasoning path as a whole, our approach applies 
support-based filtering at the slot-cluster level, retaining only 
clusters backed by enough distinct paths. This allows the framework 
to filter out hallucinated intents and noisy slot predictions at a 
finer granularity than path-level scoring.

\subsection{Support-Based Filtering}
\label{sec:filtering}
Based on the computed support score, we filter out 
unreliable clusters. Similar to the path-level filtering 
explored in reasoning-aware self-consistency 
frameworks~\cite{wan-etal-2025-reasoning}, we apply a 
minimum support threshold to discard clusters that lack 
sufficient cross-path agreement. Specifically, clusters 
are retained only if they satisfy:
\begin{equation}
    \mathcal{C}^* = \left\{ C \in \mathcal{C} \mid 
    \text{sup}_p(C) \geq \left\lceil \frac{K}{2} \right\rceil 
    \right\}
\end{equation}
The support criterion discards hallucinated intents supported 
by fewer than half the paths.

\subsection{Re-Integration} 
To re-integrate the clusters into the frames of the final prediction, 
we directly use the frame domain and intent. For slots, we propose a 
Value-First re-integration strategy inspired by Token-level Self-consistency from CroPrompt \cite{croprompt}, where representative 
slot values are first identified by support, and the corresponding slot 
keys are then determined by majority vote.

We compute the \textbf{value support score}, which counts the number 
of distinct frames containing value $v$ at least once. Then we filter out the values less than half frames support.
\begin{equation}
    \mathrm{sup}_V(v) = \left| \left\{ F \in C \mid \exists\, k : 
    (k, v) \in F \right\} \right|
\end{equation}
\begin{equation}
    V^* = \left\{ v \mid \mathrm{sup}_V(v) \geq \left\lceil \frac{|C|}{2} \right\rceil \right\}
\end{equation}

In the second step, we determine the corresponding slot keys by majority vote for each retained value $v \in V^*$:  
\begin{equation}
    k^*(v) = \underset{k}{\arg\max} \sum_{F \in C} \mathds{1}((k, v) \in F)
\end{equation}

A representative frame $F^*$ is then produced by fixing domain and 
intent from the bucket key, with the final slot set constructed as:
\begin{equation}
    F^* = (d,\ i,\ s^*), \quad s^* = \{ (k^*(v), v) \mid v \in V^* \}
\end{equation}
The final multi-intent prediction is:
\begin{equation}
    Y^* = \{ F^*_C \mid C \in \mathcal{C}^* \}
\end{equation}

\subsection{Handling Empty Semantics}
\label{sec:empty}
Reasoning paths may occasionally produce malformed or empty outputs. Frames with missing \texttt{domain} or \texttt{intent} fields are discarded prior to frame pool construction. If the resulting frame pool is empty, the system returns an empty prediction $Y^* = \emptyset$.

\section{Experiments}
\subsection{Experiment Setting}
\subsubsection{Datasets}
We evaluate on MAC-SLU~\cite{macslu}, 
a Chinese multi-intent SLU dataset for automotive cabin scenarios, 
spanning 8 domains, 81 intents, and 192 slot types with up to 4 
simultaneous intents per utterance. Semantic annotations are 
structured as semantic frames, consistent with the frame-level 
design of SFL-MTSC.

\subsubsection{Models}
We evaluate three model configurations: (1) 
Qwen3-4B-Instruct~\cite{yang2025qwen3technicalreport} for 
text-based inference, (2) a pipeline system combining 
Whisper-Large-V3-Turbo~\cite{radford2023robust} (CER = 12.83\% 
on the MAC-SLU test set) for ASR with Qwen3-4B-Instruct for 
NLU, and (3) Qwen2.5-Omni-7B~\cite{xu2025qwen25omnitechnicalreport} 
as an end-to-end LALM for direct speech-to-semantics inference.

\subsubsection{Implementation Details}
All experiments were conducted on an NVIDIA Titan RTX GPU. 
For LLM inference, we employed vLLM~\cite{kwonEfficientMemoryManagement2023} 
for accelerated deployment. For Qwen2.5-Omni-7B, we used 
vLLM-Omni~\cite{yinVLLMOmniFullyDisaggregated2026}, an 
inference framework that extends vLLM with support for 
omni-modal language models.

\subsubsection{Baselines}
We compare SFL-MTSC against three prompting baselines:
Vanilla Prompting~\cite{macslu}, which directly extracts 
intents and slots in a single prompt;
CroPrompt~\cite{croprompt}, which first performs intent 
detection and then leverages the results for slot filling; 
and GPT-SLU~\cite{gpt-slu}, which jointly predicts intents 
and slots in the first stage and refines predictions via 
cross-task information exchange in the second stage.
Since CroPrompt and GPT-SLU were originally designed for 
zero-shot single-intent scenarios, we extend their prompt 
templates to support multiple semantic frames per utterance.

\subsubsection{Evaluation Metric}
We followed the standard metrics for SLU tasks \cite{gpt-slu}. For intent detection, we calculated accuracy, and for slot filling, we calculated the F1 score. The Overall Acc. measures the 
proportion of utterances for which both intent detection and slot 
filling are simultaneously correct, serving as the strictest end-to-end 
evaluation criterion.

\subsection{Main Results}
All experiments are conducted in a zero-shot setting, 
where no task-specific training or fine-tuning is performed.
The main results are shown in Table~\ref{tab:main_result}. 
For each prompting method, we generate K = 5 reasoning paths at sampling temperatures 
$T \in \{0, 0.3, 0.5, 0.7, 1.0\}$. Each path produces an 
independent set of semantic frame predictions for the same input 
utterance. The five predictions are then aggregated by SFL-MTSC, 
which performs frame-level clustering, path support scoring, 
filtering, and re-integration to produce the final multi-intent 
prediction $Y^*$. In this work, we set the hyperparameters as 
follows: similarity threshold $\tau = 0.55$ and hybrid Jaccard 
mixing coefficient $\alpha = 0.3$, which balances Overall Accuracy and Slot F1 performance.

Our observations are as follows: 

\noindent\textbf{(1) SFL-MTSC consistently improves Overall Acc. 
across all zero-shot settings,} demonstrating the effectiveness of 
frame-level aggregation as a general post-inference strategy that 
requires no additional supervision or fine-tuning.

\noindent\textbf{(2) SFL-MTSC yields the largest gains under 
zero-shot Vanilla Prompting.} We observe improvements of 1.23\%, 
0.4\%, and 1.45\% in Overall Acc. for the three respective models. 
This is because Vanilla Prompting provides the least structural 
guidance, leading to higher cross-path variability in zero-shot 
decoding, which SFL-MTSC can effectively resolve through frame-level 
aggregation.

\noindent\textbf{(3) Gains diminish for prompting methods with 
stronger structural guidance.} For CroPrompt and GPT-SLU, which 
impose more structured zero-shot prompting strategies, Overall Acc. 
gains are more modest, indicating that frame-level filtering may 
over-regularize already stable predictions.

\noindent\textbf{(4) SFL-MTSC regularizes slot structure more 
effectively than intent classification.} Across all zero-shot 
configurations, Slot F1 shows more consistent improvements than 
Intent Acc., which fluctuates within 1--7\% and occasionally 
decreases slightly. This suggests that in zero-shot inference, 
cross-path inconsistency primarily manifests at the slot level, 
while intent predictions tend to be relatively stable across 
reasoning paths even without aggregation.

\begin{table}[t]
    \centering
    \resizebox{\columnwidth}{!}{%
    \begin{tabular}{cc|lll}
        \toprule
        Intent Filter & Slot Filter & Overall Acc. & Intent Acc. & Slot F1 \\
        \midrule
        \ding{55} & \ding{55} & 2.52 & 31.19 & 58.13\\
        \ding{51} & \ding{55} & 2.69 & 32.23 & 58.18\\
        \ding{55} & \ding{51} & \textbf{3.30} & \textbf{44.22} & \textbf{58.25}\\
        \ding{51} & \ding{51} & \textbf{3.30} & \textbf{44.22} & \textbf{58.25}\\
        \bottomrule
    \end{tabular}%
    }
    \caption{Ablation results on support-based filtering configurations. All experiments use Qwen3-4B-Instruct-2507~\cite{yang2025qwen3technicalreport} with vanilla prompting~\cite{macslu}. The best results are shown in \textbf{bold}.}
    \label{tab:ablation_support}
\end{table}

\subsection{Effect of Support-Based Filtering}
To examine the effect of support-based filtering, we conduct an 
ablation study varying the placement of the support filter across 
two levels: domain--intent and slot cluster. We evaluate four 
configurations: no filtering, intent-level only, slot-level only, 
and both levels combined, and intent-level filtering is installed between domain-intent clustering and slot clustering. All experiments are conducted using 
Qwen3-4B-Instruct-2507 with vanilla prompting. 

The results are illustrated in Table~\ref{tab:ablation_support}. 
We observe that the slot-level filter alone is sufficient to achieve 
the highest Overall Acc. of 3.30\% and Intent Acc. of 44.22\%, with 
the combined configuration leads to identical results, indicating 
that the slot-level filter is the primary driver of performance gains, 
while adding intent-level filtering alone contributes only marginal 
improvement. Furthermore, removing all filtering leads to a 
substantial drop of 13.03\% in Intent Acc., demonstrating the 
necessity of support-based filtering for reliable multi-intent 
prediction.

\begin{table}[t]
    \centering
    \resizebox{\columnwidth}{!}{%
    \begin{tabular}{l|lllllll}
        \toprule
        $\alpha$ & 0 & 0.1 & 0.3 & 0.5 & 0.7 & 0.9 & 1 \\
        \midrule
        Overall Acc. & \textbf{3.30} & \textbf{3.30} & \textbf{3.30} & \textbf{3.30} & \textbf{3.30} & 3.21 & 3.13\\
        Intent Acc.  & 43.61 & 43.61 & 44.22 & \textbf{44.48} & 44.20 & 43.87 & 43.53\\
        Slot F1  & 58.38 & \textbf{58.47} & 58.25 & 58.23 & 57.57 & 56.59 & 54.71\\
        \bottomrule
    \end{tabular}%
    }
    \caption{The results on using different Hybrid Jaccard mixing coefficient. All experiments use Qwen3-4B-Instruct-2507~\cite{yang2025qwen3technicalreport} with vanilla prompting~\cite{macslu}. The best results are shown in \textbf{bold}.}
    \label{tab:ablation_alpha}
\end{table}

\subsection{Different Hybrid Jaccard Mixing Coefficient}
In slot clustering, we use Hybrid Jaccard with mixing coefficient $\alpha$ to interpolate between Key--Value Jaccard $\mathrm{sim}_{kv}$ and Value-Based Jaccard $\mathrm{sim}_{val}$ (see Equation~(6)). To examine the effect of $\alpha$, we evaluate a range of values $\alpha$ on the MAC-SLU test set using Qwen3-4B-Instruct-2507 with vanilla prompting.

The results are shown in Table~\ref{tab:ablation_alpha}. We observe that Overall Acc. remains 
stable at 3.30\% across a wide range of $\alpha \in [0.0, 0.7]$, and 
only begins to decline at $\alpha = 0.9$, indicating that Overall 
Accuracy is largely insensitive to the mixing coefficient. For Slot F1, the best performance of 58.47\% is achieved at $\alpha = 
0.1$, with a decrease as $\alpha$ increases. This suggests 
that slot clustering benefits more from value-based similarity than 
from key-value-based similarity.

Based on these observations, we select $\alpha = 0.3$ as the default 
setting, which maintains peak Overall Acc. while preserving 
competitive Intent Acc. and Slot F1 performance.

\section{Conclusion}
We proposed SFL-MTSC, a semantic frame-level self-consistency 
framework for robust multi-intent SLU. It decomposes predictions 
into intent-specific frames, applies Hybrid Jaccard slot clustering, 
and filters unreliable frames via path support scoring. Zero-shot 
experiments on MAC-SLU show consistent improvements in Overall Acc.\ 
and Slot F1, with the largest gains under Vanilla Prompting (Slot F1 
up to +28.86\%). Limitations include occasional drops in Intent 
Accuracy, limited gains in LALM settings due to high decoding 
variance, and evaluation on a single dataset. Future work will 
explore finer-grained intent clustering and aggregation mechanisms 
better suited for LALMs.

\newpage

\section{Acknowledgments}
This work was supported in part by Realtek Semiconductor Corporation under Grant Numbers 113KK01103 and 114KK01005. Any findings and implications in the paper do not necessarily reflect those of the sponsors.

\section{Generative AI Use Disclosure}
We used Claude Sonnet 4.6 to assist with editing and 
polishing the manuscript. We also used AI tools to assist 
with code development, with strict human review and 
verification to ensure correctness.

\bibliographystyle{IEEEtran}
\bibliography{mybib}

\end{document}